\documentclass[conference]{IEEEtran}
\IEEEoverridecommandlockouts
\usepackage{cite}
\usepackage{float}
\usepackage{amsmath,amssymb,amsfonts}
\usepackage{graphicx}
\usepackage{textcomp}
\usepackage{algorithm}
\usepackage{algpseudocode}
\usepackage{xcolor}
\def\BibTeX{{\rm B\kern-.05em{\sc i\kern-.025em b}\kern-.08em
    T\kern-.1667em\lower.7ex\hbox{E}\kern-.125emX}}
\begin{document}

\title{Hierarchical Sampling-based Planner with LTL Constraints and Text Prompting\\
}

\author{\IEEEauthorblockN{1\textsuperscript{st} Jingzhan Ge}
\IEEEauthorblockA{\textit{Dept. of Electrical and Computer Eng.} \\
\textit{University of Florida}\\
Gainesville, FL \\
jingzhange@ufl.edu}
\and
\IEEEauthorblockN{2\textsuperscript{nd} Zi-Hao Zhang}
\IEEEauthorblockA{\textit{Dept. of Mechanical and Aerospace Eng.} \\
\textit{University of Florida}\\
Gainesville, FL \\
zhangzihao@ufl.edu}
\and
\IEEEauthorblockN{3\textsuperscript{rd} Sheng-En Huang}
\IEEEauthorblockA{\textit{Dept. of Electrical and Computer Eng.} \\
\textit{University of Florida}\\
Gainesville, FL \\
huang.sh@ufl.edu}
}

\maketitle

\begin{abstract}
This project introduces a hierarchical planner integrating Linear Temporal Logic (LTL) constraints with natural language prompting for robot motion planning. The framework decomposes maps into regions, generates directed graphs, and converts them into transition systems for high-level planning. Text instructions are translated into LTL formulas and converted to Deterministic Finite Automata (DFA) for sequential goal-reaching tasks while adhering to safety constraints. High-level plans, derived via Breadth-First Search (BFS), guide low-level planners like Exploring Random Trees (RRT) and Probabilistic Roadmaps (PRM) for obstacle-avoidant navigation alog with LTL tasks. The approach demonstrates adaptability to various task complexities, though challenges such as graph construction overhead and suboptimal path generation remain. Future directions include extending to considering terrain conditions and incorporating higher-order dynamics.
\end{abstract}

\begin{IEEEkeywords}
LTL, Motion Planning, Kinematic, RRT, PRM.
\end{IEEEkeywords}

\section{Introduction}
Motion and path planning are fundamental challenges in robotics, enabling autonomous systems to navigate complex environments effectively. Traditional approaches to motion planning often rely on algorithms like Rapidly Exploring Random Trees (RRT~\cite{karaman2011anytime}) and Probabilistic Roadmaps (PRM~\cite{kavraki1996probabilistic}), which are well known for their efficiency in exploring high-dimensional spaces. However, these conventional methods do not inherently consider complex temporal and logical task constraints, limiting their ability to handle more sophisticated behaviors required in dynamic or structured environments.

In this paper, we present a novel approach: a Hierarchical Sampling-based Planner with Linear Temporal Logic (LTL) constraints. By integrating LTL, our approach can specify and satisfy complex temporal goals (like conjunctive-disjunctive sequential task), allowing more versatile path planning that respects high-level task requirements. Our hierarchical framework combines the benefits of sampling-based planning with the expressiveness of temporal logic, addressing scenarios that are challenging for conventional RRT or PRM alone. This work contributes to bridging the gap between low-level motion planning and high-level task specification, enhancing the capabilities of autonomous systems in structured and dynamic settings.

In our approach, we need to perform cell decomposition and high-level planning identify the regions that will be used for low-level planning instead of considering the entire map, effectively guiding the low-level planner and improving overall efficiency. Our approach aims to provide a more flexible and reliable solution for scenarios where traditional sampling-based methods fall short, especially in environments where adherence to sequential goals or safety conditions is essential. Using a hierarchical structure, we improve both the scalability and efficiency of the planner, making it suitable for a wide range of robotic applications.

\section{Preliminary}
\subsection{Sampling-based Motion Planner}

Sampling-based motion planners are efficient algorithms designed to explore high-dimensional state spaces by randomly sampling points and connecting them to construct obstacle-free paths~\cite{karaman2011anytime}. Unlike traditional grid-based methods, these algorithms do not require an exhaustive exploration of the entire configuration space, making them faster and more scalable for complex environments. The number of iterations and sampling density directly influence the connectivity of the graph and the optimality of the generated path~\cite{marinplaza2018optimal}.

One notable limitation of these planners is their difficulty in traversing narrow passages or tight spaces, where random sampling often fails to establish sufficient connectivity. Despite this, sampling-based methods remain popular for their simplicity and ability to handle high-dimensional problems effectively.

Two widely used algorithms in this category are Probabilistic Roadmap (PRM) and Rapidly-exploring Random Trees (RRT). These algorithms differ primarily in their approach to sampling and connecting points to construct the graph. Optimized variants, such as PRM* and RRT*, improve upon their respective base algorithms by incorporating cost-based optimization, which enhances path quality. We provide detailed explanations of PRM and RRT in subsequent sections of this paper, highlighting their mechanics and use cases.

\subsection{Language Model}
This study presents a comprehensive framework for fine-tuning an autoregressive language model, Llama 3.2-1B Instruct~\cite{touvron2023llama}, to translate natural language descriptions into Linear Temporal Logic (LTL) formulas. The workflow includes data preparation by pairing prompts with their corresponding LTL formulas, tokenizing and encoding them to ensure model compatibility. A custom dataset class is developed to align inputs and labels while excluding padding tokens from loss computation. The training process employs Hugging Face’s Trainer and DataCollatorForSeq2Seq, enabling efficient dynamic padding and optimization with configurable hyperparameters. Post-training, the model is evaluated using the text-generation pipeline to validate its capability in mapping natural language prompts to formal logic. This approach offers a scalable solution for translating human instructions into formal specifications, with potential applications in automated planning, formal verification, and control systems.
\begin{figure}
    \centering
    \includegraphics[width=0.8\linewidth]{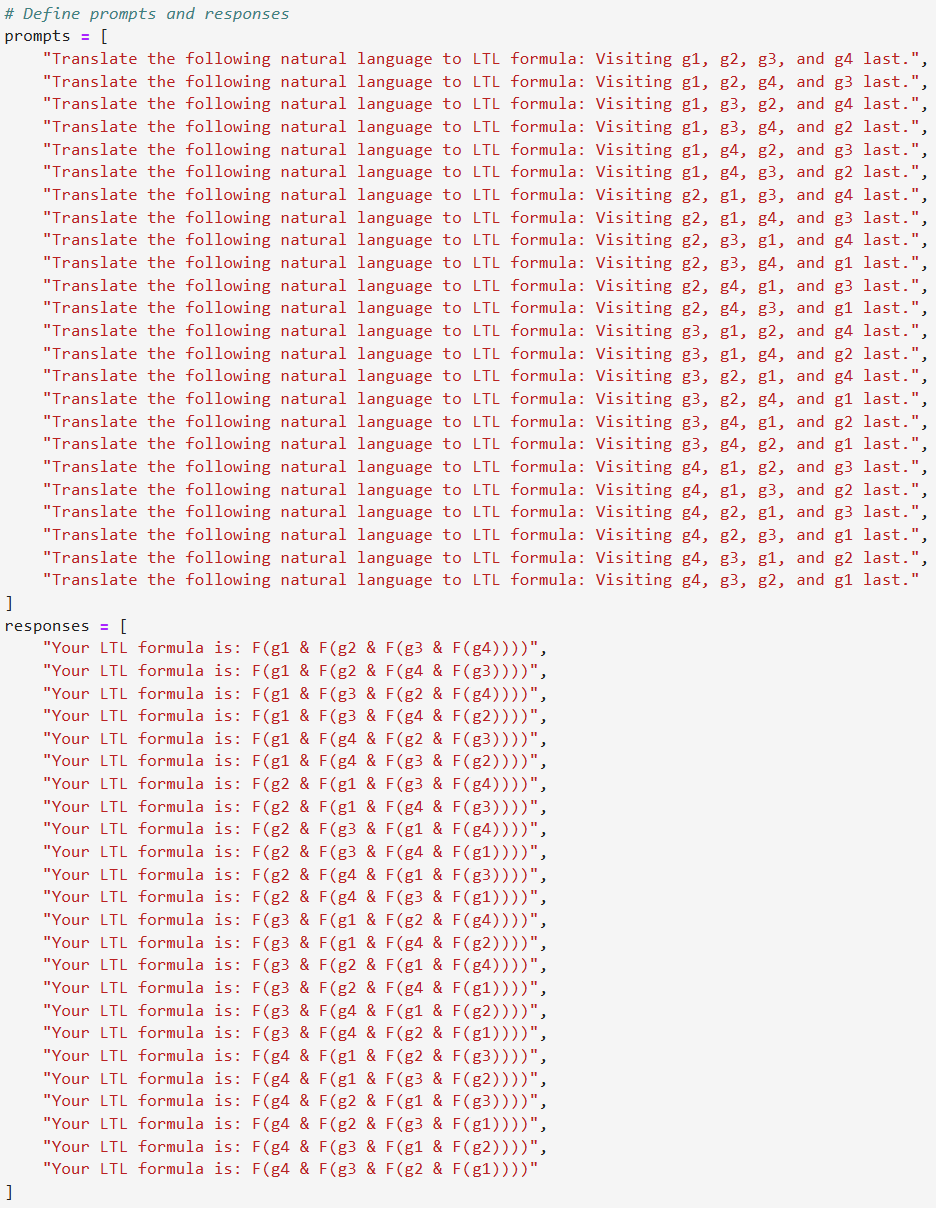}
    \caption{The training data for the conversion from natural language to LTL formula.}
    \label{fig:nl2ltl_training}
\end{figure}

\section{Methodology}
\subsection{Overview}
Our approach can be divided into four main components: natural language to LTL conversion, high-level planning, low-level planning, and Robot Kinematic Model verification. First, natural language commands are translated into LTL formulas, which may specify a sequential task order such as "visit goal1, then visit goal2, then visit goal3, and finally visit goal4" in strict order. This transformation allows us to formally represent complex tasks in a manner that can be systematically addressed by our planner.

Next, we proceed to the core of our method: high-level planning. Using cell decomposition, we derive a graph that describes the transitions between each decomposed cell, which is subsequently converted into a Transition System (TS). Combined with the LTL formula, which is transformed into a Deterministic Finite Automaton (DFA) to represent temporal logic constraints, we compute the product transition system that integrates both the task and environmental (map). Subsequently, we employ breadth-first search (BFS) to determine the high-level planning route, identifying the sequence of decomposed regions that lead to the goal. These regions serve as a guide for low-level planning.

Based on the high-level planning results, we then perform sampling-based low-level planning exclusively within the regions identified by the high-level plan to generate a detailed trajectory. This targeted approach significantly enhances the efficiency of the low-level planner by reducing the search space.

Finally, the generated trajectory is verified using a differential drive robot and a PID controller to ensure the feasibility of the trajectory by executing and following it. This verification step ensures that the planned trajectory can be realistically executed by a physical robot.

\subsection{Natural Language to LTL}
In this paper, we leverage Meta's large language model to achieve our transformation goals. We design prompt and answer pairs as shown in Fig.~\ref{fig:nl2ltl_training} However, semantic interpretation by the large language model sometimes produces results that do not align with our expectations. To address this issue, we explore fine-tuning the model to improve its performance. The large language model utilized in our study is Llama 3.2-1B In-struct~\cite{touvron2023llama}. Despite benefiting from the capabilities of the large language model, the output results remain suboptimal, as discussed in detail in subsequent sections.

\subsection{Map Setup and Cell Decomposition}
\begin{figure}[htbp]
    \centering
    \includegraphics[width=0.8\columnwidth]{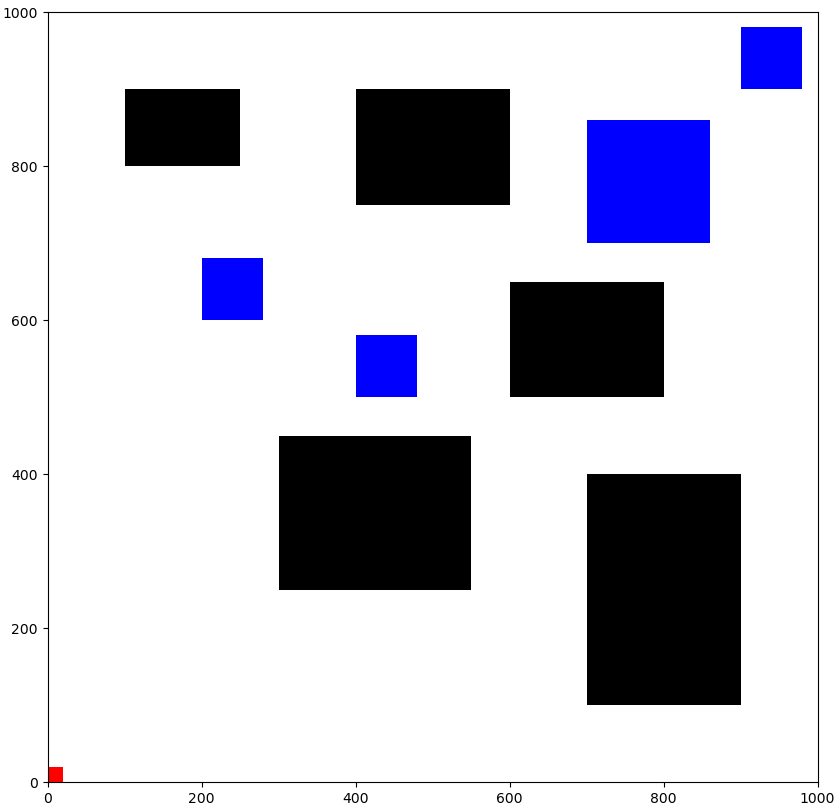}
    \caption{Map Workspace}
    \label{fig:Map Workspace}
\end{figure}

In this paper, we define the map with a start region (red), four goal regions (blue) and five obstacles (black), shown in Fig.~\ref{fig:Map Workspace}. These regions are defined by coordinate, easy to edit and add more regions. 
\begin{figure*}[htbp]
\centering
\vspace{-3mm}
\includegraphics[width=0.98\textwidth]{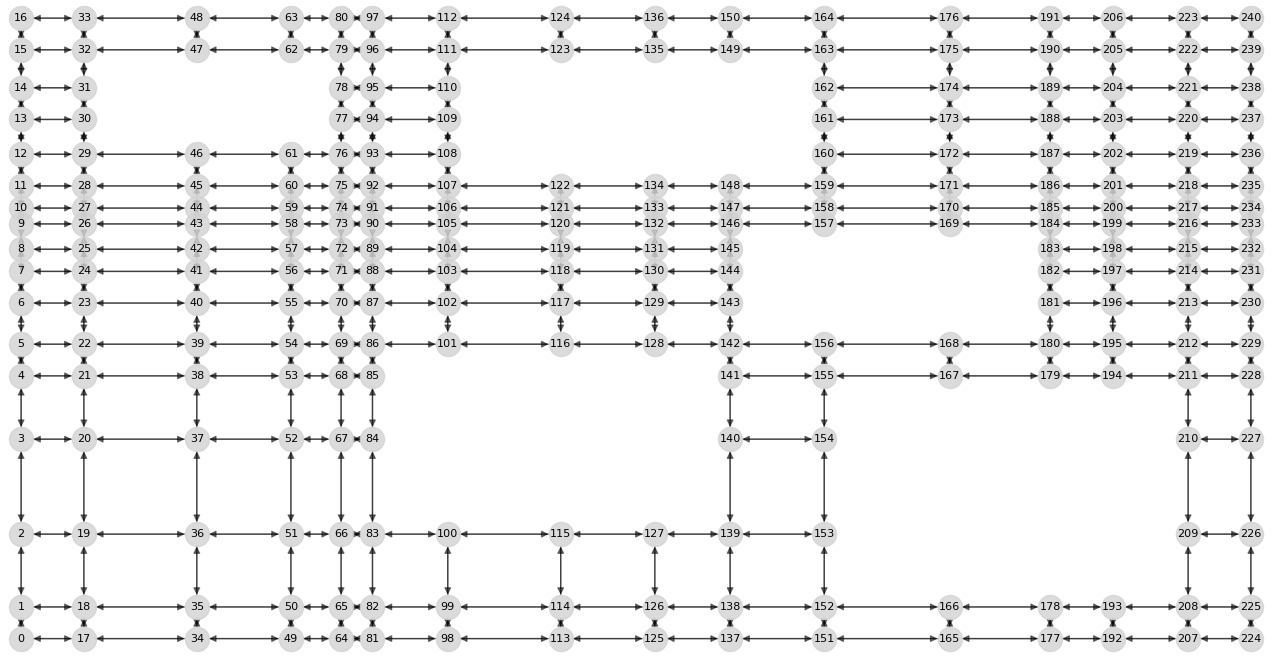}
\vspace{-2mm}
\caption{Graph}
\label{fig:Graph}
\vspace{-1mm}
\end{figure*}

\begin{figure}[htbp]
    \centering
    \includegraphics[width=0.8\columnwidth]{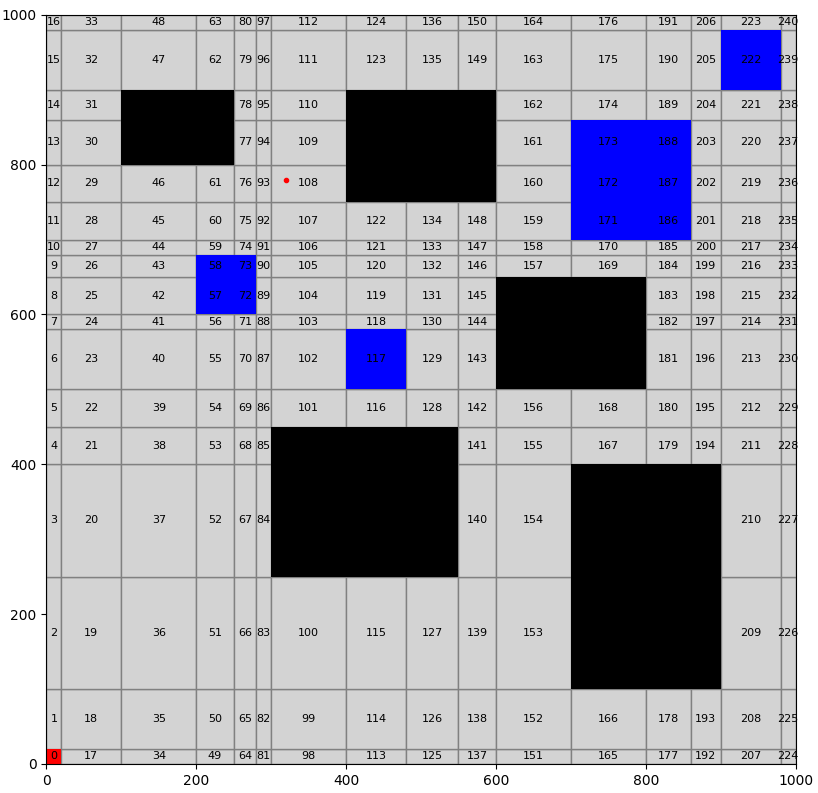}
    \caption{Cell Decomposition}
    \label{fig:Cell Decomposition}
\end{figure}

To facilitate effective planning, we utilize cell decomposition to partition the environment into manageable regions. We use a rectangle-based decomposition method that divides the environment into rectangular cells, shown in Fig.~\ref{fig:Cell Decomposition}. Each decomposed region is assigned a unique ID, simplifying subsequent handling in the transition system. These IDs are also used in low-level planning to manage sample points within each region. 

The purpose of cell decomposition is to transform a complex environment into a graph-like structure for easier analysis and navigation. Each decomposed region can only move to neighboring regions, defined as left, right, top, and bottom, shown in Fig.~\ref{fig:Graph}. Obstacles are excluded from decomposition, resulting in cells that only represent free space, which reduces computational complexity and focuses planning on feasible areas.
Excluding obstacles also ensures that the high-level plan is inherently collision-free, reducing the need for additional collision-checks and improving the overall efficiency and reliability of the planning process. But we still can add avoid regions after the decomposition which will be talked in later part.

% \begin{figure}[htbp]
%     \centering
%     \includegraphics[width=0.9\columnwidth]{Figure/grapha.png}
%     \caption{Graph}
%     \label{fig:Graph}
% \end{figure}

\subsection{High-Level Planning}
High-level planning operates on a simplified, abstract representation of the environment, which consists of interconnected cells representing the free space. This abstract representation, in the form of a Transition System (TS), allows us to efficiently determine the sequence of regions that lead to the goal while satisfying the given LTL constraints. By focusing on a high-level abstraction, the complexity of planning is significantly reduced compared to considering the entire configuration space.

High-level planning thus serves as a bridge between high-level task specifications and low-level trajectory generation, providing a structured plan that guides the sampling-based planner to operate within specific regions, ultimately improving efficiency and ensuring adherence to the desired task requirements.

\subsubsection{Transition System Generation}
The transition system is converted from the graph that we generated to describe the transition between each cell (decomposed regions). Each graph node is the transition system state which is decomposed region's ID. The edges of the graph are the transitions in the transition system. Here we didn't assign action labels to the transitions. As our graph is a directed graph, we can further define the transition as one-way from certain cell for more complex map workspace or more complex LTL tasks.
\subsubsection{LTL to DFA}
Linear Temporal Logic (LTL) is used to specify temporal properties and high-level tasks for a system. However, automating the execution of these LTL specifications requires converting them into a form that can be used by algorithms for planning and verification. This is where Deterministic Finite Automata (DFA) come in. Converting LTL to DFA allows us to represent these temporal properties in a state-based model that can be easily integrated into planning processes. The DFA captures all valid sequences of states that satisfy the LTL specification, enabling systematic verification and execution of complex tasks.

For example, a co-safe LTL is
\begin{equation}
F(g1\quad\&\quad F(g2 \quad\& F(g3 \quad\& F(g4))))
\label{equ:ltl}
\end{equation}
and we developed a custom fuction that can convert the LTL into DFA. The transition of the DFA is shown in the Fig.~\ref{fig:DFA1} below. 

To convert the transition system to a DFA, we need a labeling function that marks the states of the transition system with atomic propositions. These labels are used to determine if a sequence of states satisfies the LTL specification. Here we can use labelling to handle additional obstacle regions which means we can handle co-safe condition, the results are shown in result section.

We define there are 5 DFA states, (q0 to q4) are the normal states that represent the current progress in satisfying the LTL formula. q5 is the trap state when violate the rules.

\begin{figure}[htbp]
    \centering
    \includegraphics[width=0.8\columnwidth]{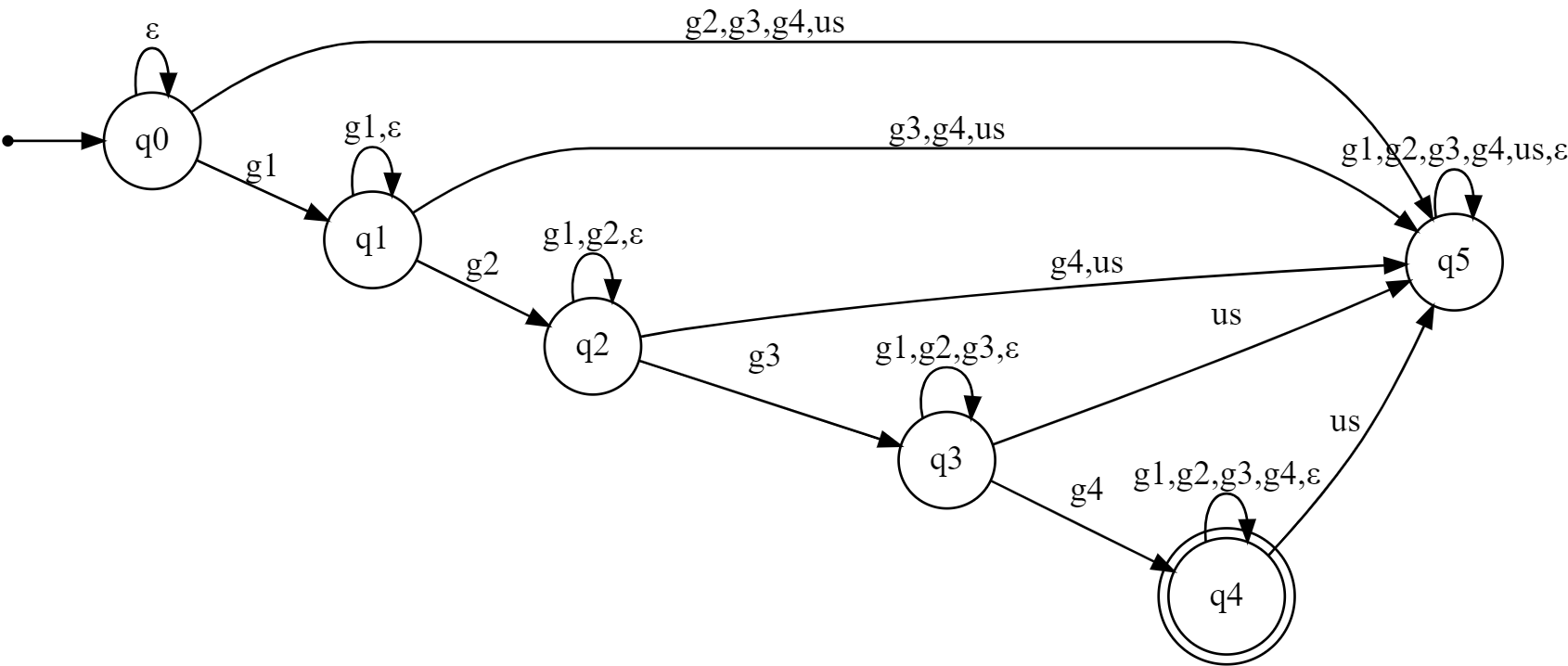}
    \caption{DFA}
    \label{fig:DFA1}
\end{figure}

\subsubsection{Product Transition System}
The Product Transition System provides a way to track both the system's position (first element in product state, transition system state) and the task's progress (second element is the DFA state) simultaneously, which is crucial for planning and verification. By combining these two aspects, the system can ensure that the path taken not only follows feasible transitions but also satisfies the desired temporal properties defined by the LTL formula. 

\subsubsection{Breadth First Search}

Based on state product, we perform the Breadth First Search (BFS). BFS is a graph traversal algorithm used to explore nodes level by level. BFS begins at the starting node and explores all its neighbors before moving to the next level of nodes. This method is particularly effective in scenarios where the shortest path is desired, and it guarantees that all nodes at the current depth are explored before proceeding to the next level. 

In the context of our algorithm, BFS is used to systematically explore all possible transitions (product states transition) from a given starting state to (1, 'q0') find the feasible paths to the final states which could be multiple product states (a single goal region could composed by several decomposed region). 

\begin{figure}[htbp]
    \centering
    \includegraphics[width=0.8\columnwidth]{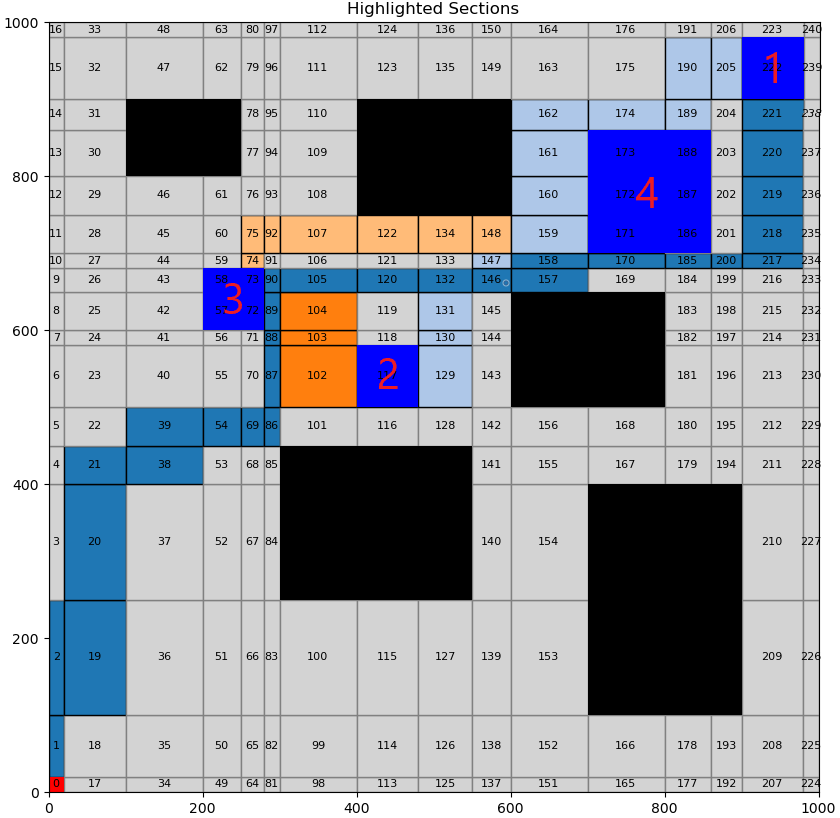}
    \caption{High-Level planning result}
    \label{fig:Figure_4a}
\end{figure}

In Fig.~\ref{fig:Figure_4a}, shows the high-level planning results for the LTL Eq.~\ref{equ:ltl}. The highlight regions are the high-level planning routes, different colors shows different stages (dark blue is for stage 1, light blue is for stage 2, dark orange is for stage 3, light orange is for stage 4, but overlapped regions only show one color). For example, the dark blue regions are the first stage that from start region (left bottom corner, red region) to the goal 1 region, we will only sample the random points in these specific regions, instead of generating random points in the whole workspace. From goal 1 to goal 2, the second stage is shown in light blue regions, again, we will only sample the random points in these specific regions. 

\subsection{Low-Level Planning (Sampling-based Planner)}
Sampling-based motion planning algorithms provide a powerful approach for navigating high-dimensional state spaces by using random sampling to construct feasible paths. Unlike traditional deterministic methods, they are well-suited for solving complex planning problems due to their ability to avoid exhaustive exploration of the configuration space. These algorithms are particularly efficient and scalable, making them a popular choice for applications where computational resources or real-time execution are critical.

Key algorithms in this category, such as Probabilistic Roadmap (PRM) and Rapidly-exploring Random Trees (RRT), offer distinct strategies for sampling and connecting points to generate obstacle-free paths. In this paper, we delve into the mechanics of these algorithms to demonstrate their effectiveness in addressing diverse planning challenges.

We modify the RRT/PRM algorithms to adapt to our cell decomposition strategy. Each random point is assigned a region ID label, making it easy to determine its position within the decomposed map. Unlike the original RRT and PRM algorithms use the collision-free function to check if the sampling points are in the obstacle regions, our approach uses cell decomposition results to preemptively exclude obstacle regions. After decomposing the map, we identify free and obstacle cells, restricting sampling and transitions to non-obstacle regions only, which avoid a traversing explicitly marked obstacles. This ensures RRT and PRM avoid generating points in obstacle regions, effectively achieving collision-free functionality while improving planning efficiency. Moreover, this method enforces transition rules, allowing transition only within the current cell or neighboring regions, ensuring the final trajectory adheres to these constraints. By integrating cell decomposition, we enhance RRT and PRM's ability to handle safety constraints and goal scheduling, even without a high-level planner.

The above discussion addresses obstacles identified before cell decomposition (static obstacles). After decomposition, additional avoid regions can still be added, which will be discussed in the Simulation section. A high-level planner manages this using labeling functions, and without a high-level planner, goal scheduling can be handled through excluding unwanted regions. The core idea is that both pre-decomposition and post-decomposition obstacles are handled by prohibiting sampling points or edges generate in those regions.

\subsubsection{Rapidly-Exploring Random Tree (RRT)}
The RRT algorithm is shown in Algorithm 1.

\begin{algorithm}[H]
\label{alg:RRT}
\caption{Rapidly-exploring Random Tree (RRT)}
\begin{algorithmic}[1]
\Require Start state $x_{start}$, goal state $x_{goal}$, search space $\mathcal{X}$, maximum iterations $K$, step size $\Delta q$
\Ensure A path from $x_{start}$ to $x_{goal}$, if found

\State $T \gets \{x_{start}\}$ \Comment{Initialize the tree with the start state}
\For{$k = 1$ to $K$}
    \State $x_{rand} \gets$ SampleRandomState($\mathcal{X}$)
    \State $x_{nearest} \gets$ NearestNeighbor($T, x_{rand}$)
    \State $x_{new} \gets$ Steer($x_{nearest}, x_{rand}, \Delta q$)
    \If{CollisionFree($x_{nearest}, x_{new}$)}
        \State Add $x_{new}$ to $T$
        \If{$x_{new}$ is near $x_{goal}$ and CollisionFree($x_{new}, x_{goal}$)}
            \State Add $x_{goal}$ to $T$
            \State \Return Path($T, x_{start}, x_{goal}$)
        \EndIf
    \EndIf
\EndFor
\State \Return Failure
\end{algorithmic}
\end{algorithm}

\subsubsection{Probabilistic Roadmap (PRM)}
 The PRM algorithm is shown in Algorithm. 2.
\begin{algorithm}
\label{alg:PRM}
\caption{Probabilistic Roadmap (PRM)}
\begin{algorithmic}[1]
\Require Search space $\mathcal{X}$, number of nodes $N$, number of neighbors $k$
\Ensure A roadmap graph $G = (V, E)$

\State $V \gets \emptyset$; $E \gets \emptyset$ \Comment{Initialize the roadmap graph}
\For{$i = 1$ to $N$}
    \State $x_{rand} \gets$ SampleRandomState($\mathcal{X}$)
    \If{CollisionFree($x_{rand}$)}
        \State Add $x_{rand}$ to $V$
        \State $N_{neighbors} \gets$ NearestNeighbors($x_{rand}, V, k$)
        \For{each $x_{near} \in N_{neighbors}$}
            \If{CollisionFree($x_{rand}, x_{near}$)}
                \State Add edge $(x_{rand}, x_{near})$ to $E$
            \EndIf
        \EndFor
    \EndIf
\EndFor
\State \Return $G = (V, E)$
\end{algorithmic}
\end{algorithm}
\subsection{Robot Kinematic Model}

The dynamics of the differential drive robot are essential to verify the feasibility of the planned trajectories. The robot's motion is governed by the following kinematic and control equations:

\subsubsection{Kinematic Model}

1. \textbf{Linear velocity:}
\begin{equation}
v = \frac{v_{\text{left}} + v_{\text{right}}}{2}
\end{equation}

2. \textbf{Angular velocity:}
\begin{equation}
\omega = \frac{v_{\text{right}} - v_{\text{left}}}{L}
\end{equation}

3. \textbf{Position update equations:}
\begin{align}
x_{\text{new}} &= x_{\text{current}} + v \cdot \cos(\theta) \cdot \Delta t \\
y_{\text{new}} &= y_{\text{current}} + v \cdot \sin(\theta) \cdot \Delta t \\
\theta_{\text{new}} &= \theta_{\text{current}} + \omega \cdot \Delta t
\end{align}

4. \textbf{Orientation normalization:}
\begin{equation}
\theta = (\theta + \pi) \bmod (2 \pi) - \pi
\end{equation}

\subsubsection{Error Dynamics}

The robot's trajectory is evaluated based on errors between the current state and the target state. The error dynamics are computed as follows:

1. \textbf{Distance error:}
\begin{equation}
e_{\text{distance}} = \sqrt{(x_{\text{target}} - x)^2 + (y_{\text{target}} - y)^2}
\end{equation}

2. \textbf{Desired heading:}
\begin{equation}
\theta_{\text{desired}} = \arctan2(y_{\text{target}} - y, x_{\text{target}} - x)
\end{equation}

3. \textbf{Heading error:}
\begin{equation}
e_{\text{heading}} = (\theta_{\text{desired}} - \theta + \pi) \bmod (2 \pi) - \pi
\end{equation}

\subsubsection{PID Controller}

To ensure that the robot follows the planned trajectory accurately, a PID controller adjusts the angular velocity based on the heading error:
\begin{equation}
\omega_{\text{desired}} = K_p \cdot e_{\text{heading}} + K_i \cdot \int e_{\text{heading}} \, dt + K_d \cdot \frac{d(e_{\text{heading}})}{dt}
\end{equation}

Here, \( K_p \), \( K_i \), and \( K_d \) are the proportional, integral, and derivative gains, respectively. The PID controller minimizes the heading error, ensuring that the robot tracks the trajectory with precision.

This kinematic and dynamic model enables us to evaluate and verify the planned paths, ensuring both collision-free navigation and feasibility of execution in physical scenarios.
\section{Simulations}
In the simulation, we will present two kinds of LTL tasks, Sequential Task with Safety Constrain and Conjunctive-Disjunctive Sequential Task. The first simulation demonstrate the ability for our method that can handle not only co-safe but also safety. In this simulation, using a goal scheduling with low-level planner (without high-level planner) can also achieve the task. The second simulation shows that in the complex tasks like Conjunctive-Disjunctive Sequential Task, must corporate with the high-level planner, which shows the importance of our method.

\subsection{Natural Language to LTL formula}
\begin{figure}[htbp]
    \centering
    \includegraphics[width=0.8\linewidth]{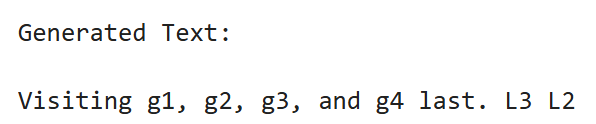}
    \caption{First result of generated text from LLaMA-3.1-8b.}
    \label{fig:nl2ltl_result}
\end{figure}

\begin{figure}[htbp]
    \centering
    \includegraphics[width=0.8\linewidth]{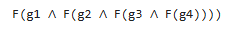}
    \caption{Second result of generated text from LLaMA-3.1-8b.}
    \label{fig:nl2ltl_result2}
\end{figure}

In this simulation, we input the prompt: "Visiting g1, g2, g3, and g4 last." The results are shown in Fig.~\ref{fig:nl2ltl_result}. It is evident that the model's output does not meet our expectations. This discrepancy can be attributed to several factors. First, the prompt design may be inadequate, as it overly constrains the model to adhere to a specific output format. Second, the dataset lacks sufficient quantity and variability. To address these issues, future research will incorporate a wider range of natural language and LTL formula pairs to enhance diversity. This approach is expected to significantly improve transformation outcomes. On the other hand, we conducted tests using Nvidia's API platform with LLaMA 3.1-8B (LLaMA-3.1-8b). By applying post-processing to the text, we obtained the results shown in Fig.~\ref{fig:nl2ltl_result2}, which align with the expected outcomes. However, further testing on additional cases is necessary to evaluate the performance of the transformation comprehensively.

\subsection{Sequential Task with Safety Constrain}
\begin{figure}[htbp]
    \centering
    \includegraphics[width=0.8\columnwidth]{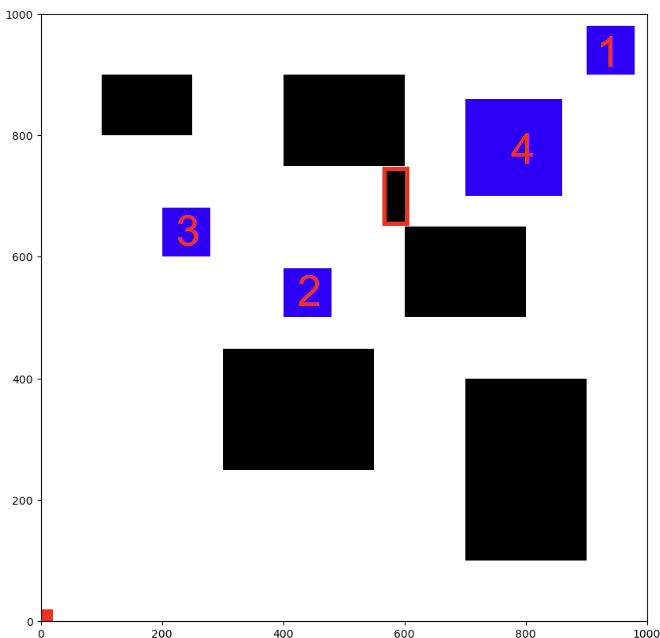}
    \caption{Sequential Task with Safety Constrain Map}
    \label{fig:map1}
\end{figure}

In this simulation, we define the task is reaching goal 1, goal 2, goal 3 and goal 4 in strict order ($g1\rightarrow g2 \rightarrow g3 \rightarrow g4$, same as previous example), and the unsafe region (in the red box) will be effect after reaching goal 1 (need detour after reaching goal 1). Shown in Fig.~\ref{fig:map1}.

\begin{figure}[htbp]
    \centering
    \includegraphics[width=0.8\columnwidth]{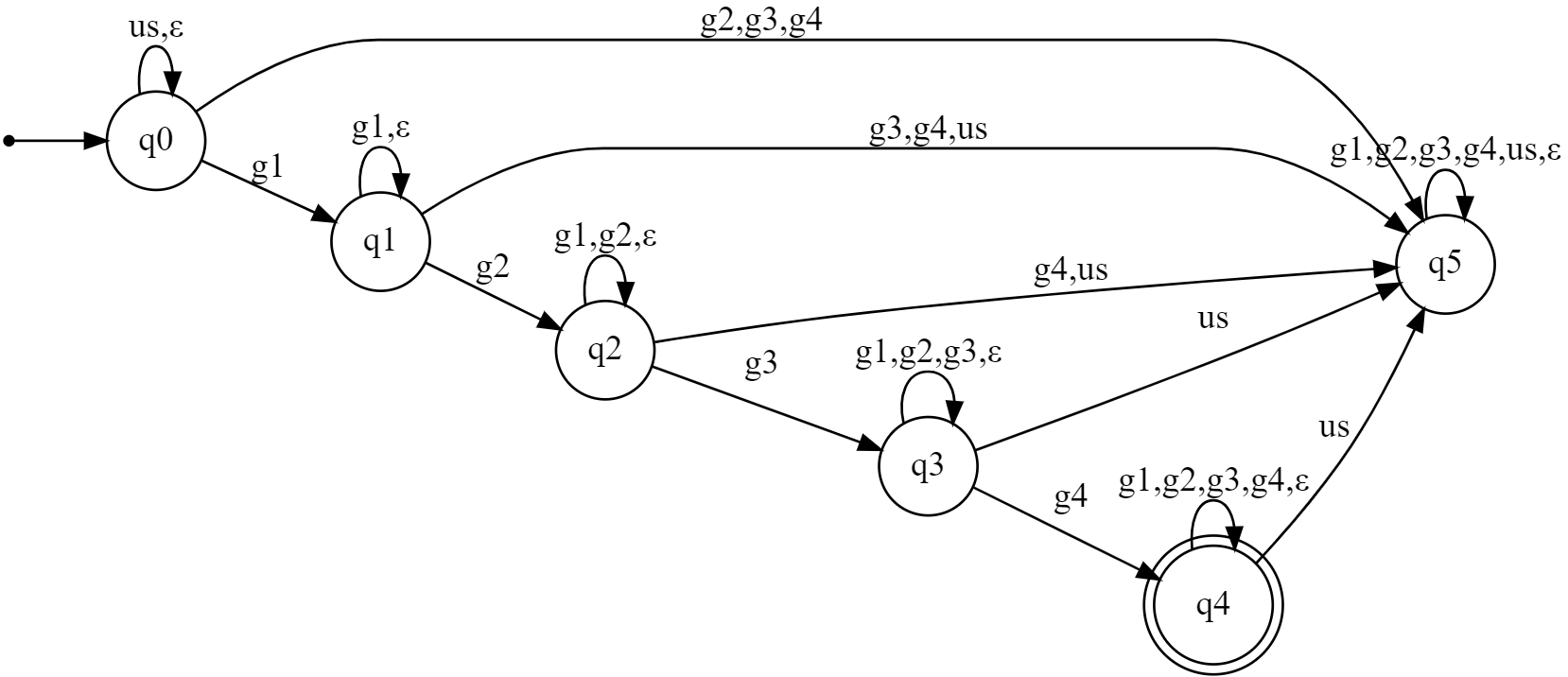}
    \caption{Sequential Task with Safety Constrain Map}
    \label{fig:DFA2}
\end{figure}

The LTL formula is
\begin{equation}
F(g1\quad\&F(g2 \quad\& F(g3 \quad\& F(g4))))\quad\&G(g1\quad\& X G(!us))
\end{equation}
And the DFA transition is shown in Fig.~\ref{fig:DFA2}

\subsubsection{Sampling-based Planner Without High-Level Planning}
\begin{figure}[htbp]
    \centering
    \includegraphics[width=0.8\columnwidth]{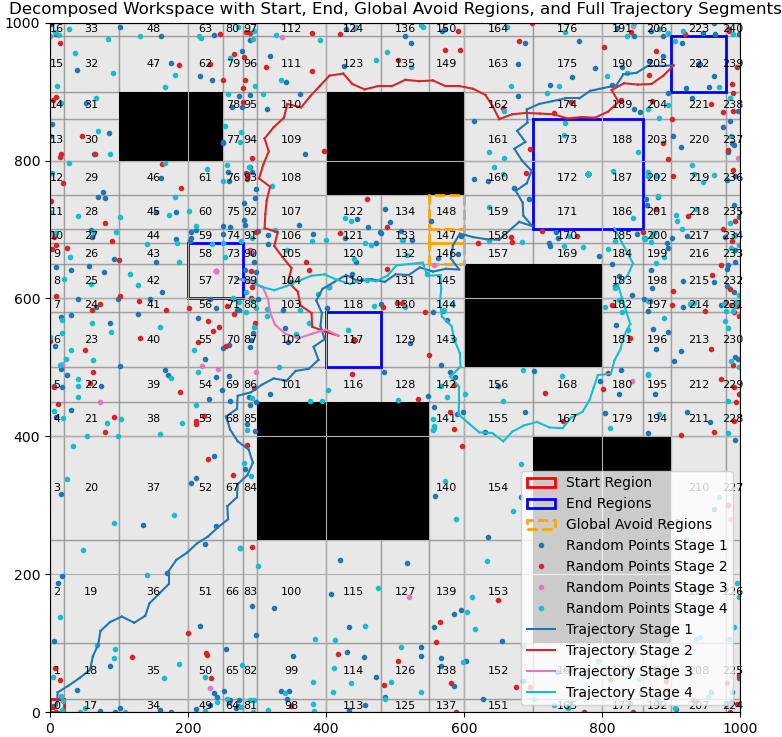}
    \caption{RRT Planner Without High-Level Planning}
    \label{fig:RRT2}
\end{figure}

In this section, we present the results using RRT/PRM with goal scheduling instead of a high-level planner. The random sample points are generated in the whole map workspace except exclude regions. The unsafe regions will be include to exclude regions according to the goal scheduling for different stages. Fig.~\ref{fig:RRT2} shows the planning results using RRT and goal scheduling without high-level planner.

\begin{figure}[htbp]
    \centering
    \includegraphics[width=0.8\columnwidth]{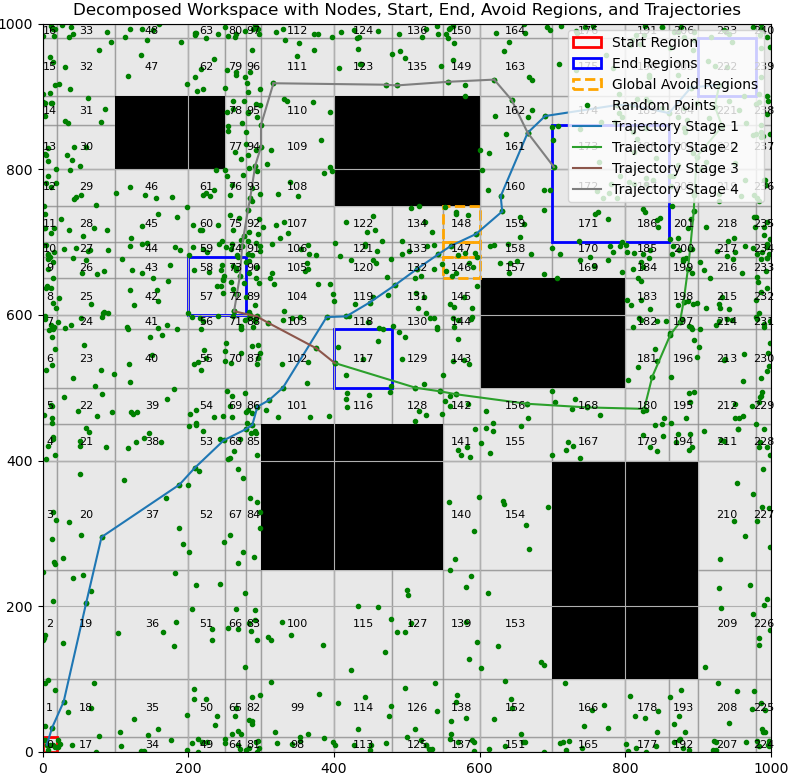}
    \caption{PRM Planner Without High-Level Planning}
    \label{fig:PRM2}
\end{figure}

Fig.~\ref{fig:PRM2} shows the planning results using RRT and goal scheduling without high-level planner.

\subsubsection{Sampling-based Planner With High-Level Planning}
\begin{figure}[htbp]
    \centering
    \includegraphics[width=0.8\columnwidth]{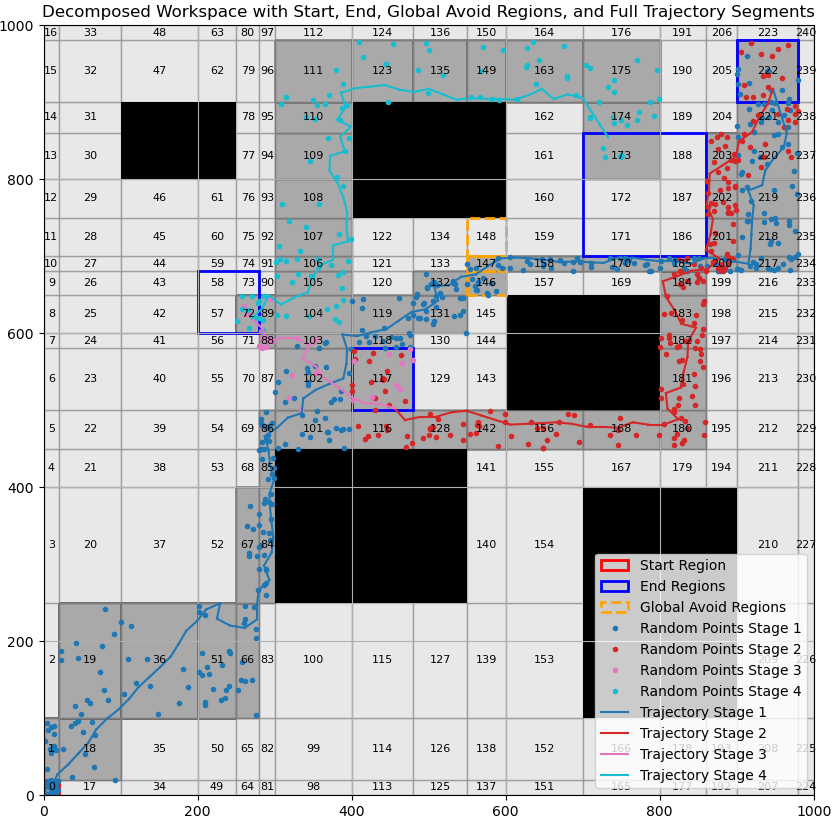}
    \caption{RRT Planner With High-Level Planning}
    \label{fig:LTL-RRT2}
\end{figure}

The highlight dark grey regions are the high level planning result. The highlighted dark grey regions are the high-level planning result.
Fig.~\ref{fig:LTL-RRT2} shows the planning results using RRT with high-level planner. 

\begin{figure}[htbp]
    \centering
    \includegraphics[width=0.8\columnwidth]{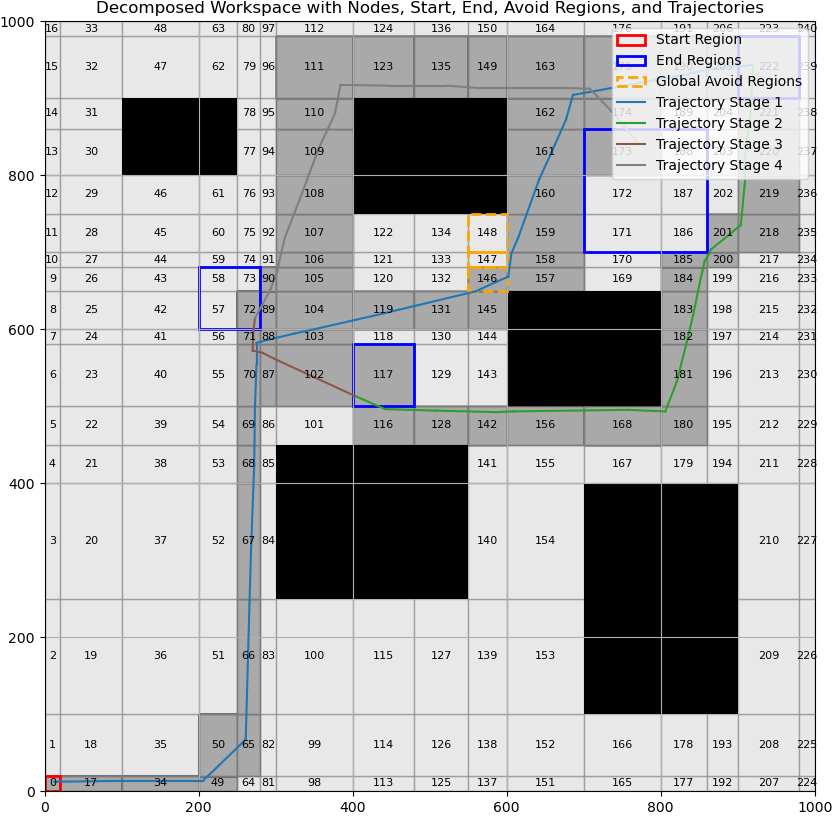}
    \caption{PRM Planner With High-Level Planning}
    \label{fig:LTL-PRM2}
\end{figure}

Fig.~\ref{fig:LTL-PRM2} shows the planning results using PRM with high-level planner.

\subsection{Conjunctive-Disjunctive Sequential Task}

\begin{figure}[htbp]
    \centering
    \includegraphics[width=0.8\columnwidth]{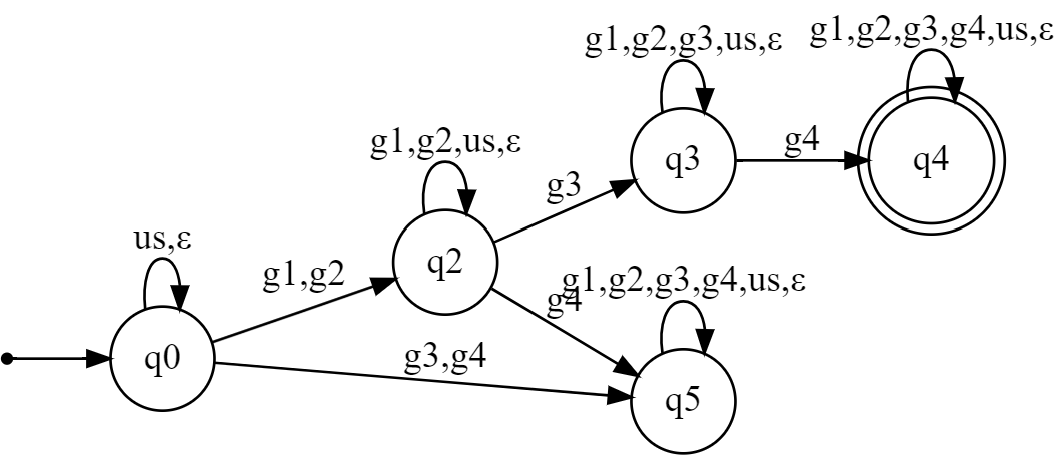}
    \caption{Sequential Task with Safety Constrain DFA}
    \label{fig:DFA3}
\end{figure}

In this simulation, we define the task is reaching goal 1 or goal 2, and reaching goal 3 and goal 4 in strict order ($g1\quad or\quad g2 \rightarrow g3 \rightarrow g4$), without other restrictions. The LTL formula is
\begin{align}
\label{equ:ltl2}
F(g1\|g2 \quad\& F(g3 \quad\& F(g4))).
\end{align}
And the DFA transition is shown in Fig.~\ref{fig:DFA3}. The map is same to previous simulation test, refer to Fig.~\ref{fig:map1}, but without the unsafe regions (red box area).

\subsubsection{Sampling-based Planner Without High-Level Planning}
In this case, this method can not handle the LTL task, as we need it reach goal 1 or goal 2 first, the goal scheduling need a clearly defined sequential task. This shows the strength of our Hierarchical Sampling-based Planner with LTL Constraints.

\subsubsection{Sampling-based Planner With High-Level Planning}
\begin{figure}[htbp]
    \centering
    \includegraphics[width=0.8\columnwidth]{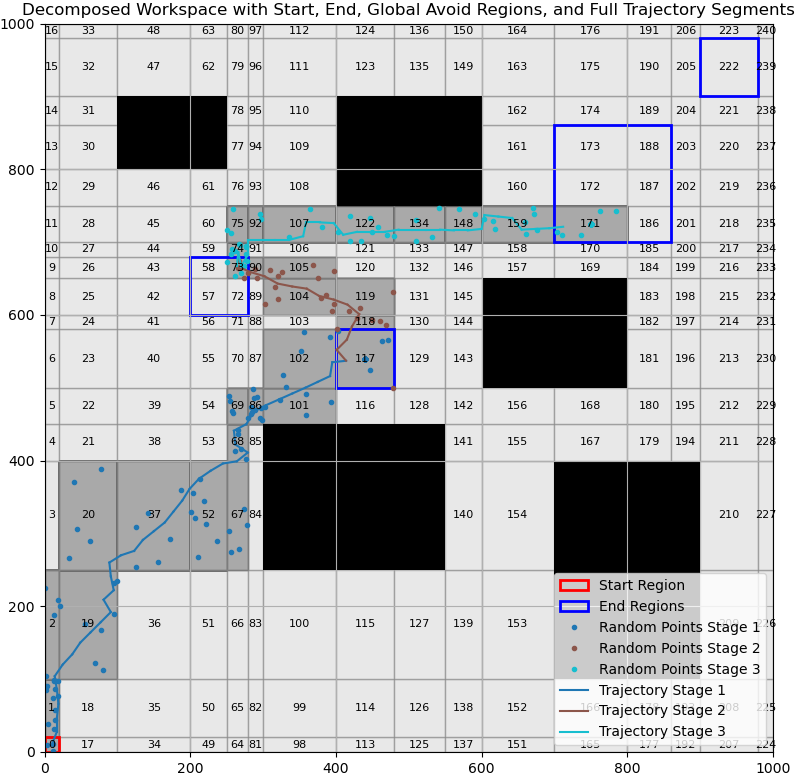}
    \caption{RRT Planner With High-Level Planning}
    \label{fig:LTL-RRT3}
\end{figure}

As the task is reaching goal 1 or goal 2 first, through the high-level planning using BFS, the algorithm will determine to visit goal 2 first only (instead of goal 1, because goal 1 is much further away from start region) and then proceed to g3, g4.
The highlighted dark grey regions are the high-level planning result.
Fig.~\ref{fig:LTL-RRT3} shows the planning results using RRT with high-level planner.

\begin{figure}[htbp]
    \centering
    \includegraphics[width=0.8\columnwidth]{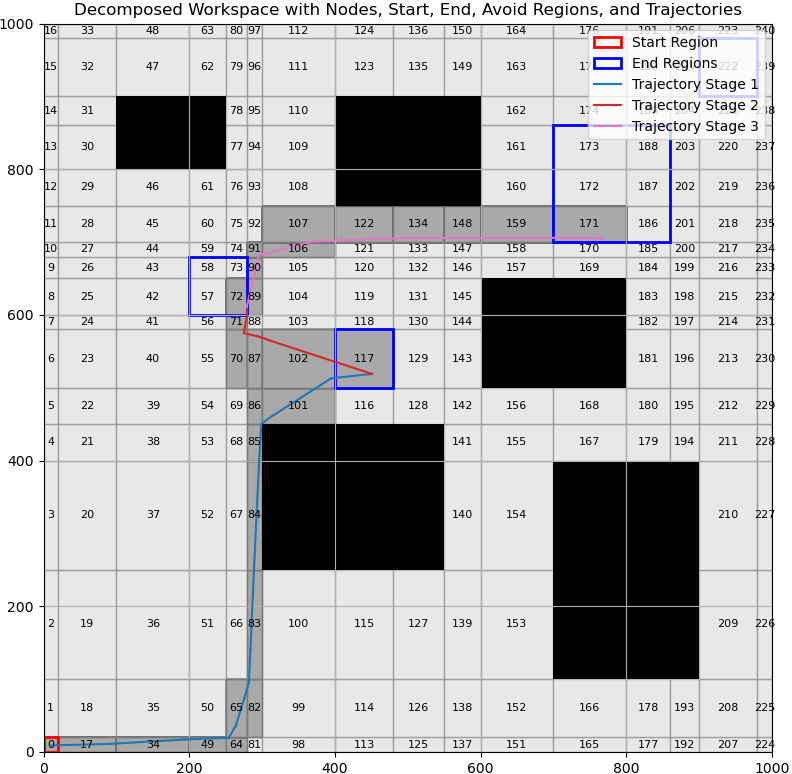}
    \caption{PRM Planner With High-Level Planning}
    \label{fig:LTL-PRM3}
\end{figure}

Fig.~\ref{fig:LTL-PRM3} shows the planning results using RRT with high-level planner.

\section{Conclusions}
In this paper, we propose an end-to-end pipeline that enables users to issue natural language commands for robot path planning and constraints. The robot translates the natural language into Linear Temporal Logic (LTL) formulas, which are then used to automatically generate corresponding Deterministic Finite Automaton (DFA) objects to detail the conversion process. Subsequently, the map is decomposed into regions, and these regions are converted into Transition System (TS), together with Deterministic Finite Automaton (DFA) to construct a product Transition System (prodTS) for high-level planning by Breadth First Search (BFS). A sampling-based method is then employed for low-level planning, integrating the robot's kinematics to achieve a complete system. The simulation results demonstrate excellent performance of our hierarchical planner in path planning and show the strength that can handle more complex tasks (e.g., reaching A or B). As the high-level planning uses the graph that describe the whole transition between each cell, we can handle more complex decomposition like triangle based decomposition more efficient. In the case of complex map workspace, the high-level planning is much more beneficial to the RRT, without growing the tree towards dead-end regions. But disadvantage of our method is that it can't give the optimal route, this can be improved when we corporate with weighted transition using Dijkstra. But it's inevitably to increase the computational cost as we compute the high-level planning. Currently, the natural language to LTL conversion did not meet expectations due to inadequate training. In future work, we will improve this component to realize a more comprehensive path-planning pipeline. Additionally with weighted transition, we can handle the task that require the consideration of terrain.

\section{Contribution}
The project was a collaborative effort, with each team member contributing to different aspects of the development. Jingzhan Ge was responsible for decomposition, graph generation, and converting the graph into a transition system. Additionally, Jingzhan Ge worked on integrating the product state, implementing the breadth-first search (BFS) algorithm for path planning and modify the low-level planner for corporation with high-level planner/goal scheduler. Zi-Hao Zhang specialized in low-level planning, implementing Rapidly-Exploring Random Trees (RRT) and Probabilistic Roadmaps (PRM). Furthermore, Zi-Hao Zhang developed the robot kinematic model, laying the mathematical foundation for motion execution. Sheng-En Huang focused on high-level planning, implementing the conversion from natural language to Linear Temporal Logic (LTL) and transforming LTL into a Deterministic Finite Automaton (DFA). Sheng-En Huang also contributed to discussions and the overall design of the motion planner architecture. Together, these contributions enabled the successful completion of the project.

\bibliographystyle{IEEEtran}
\bibliography{reference.bib}

\vspace{12pt}

\end{document}